\documentclass[preprint,1p,12pt]{elsarticle}

\makeatletter
\def\ps@pprintTitle{
  \let\@oddhead\@empty
  \let\@evenhead\@empty
  \let\@evenfoot\@oddfoot
}
\makeatother

\usepackage{amssymb,amsmath,color}
\usepackage{graphicx,enumitem}

\usepackage{url}
\urldef{\mailsa}\path|{alfred.hofmann, ursula.barth, ingrid.haas, frank.holzwarth,|
\urldef{\mailsb}\path|anna.kramer, leonie.kunz, christine.reiss, nicole.sator,|
\urldef{\mailsc}\path|erika.siebert-cole, peter.strasser, lncs}@springer.com|

\newtheorem{theorem}{Theorem}
\newtheorem{corollary}[theorem]{Corollary}

\newdefinition{definition}[theorem]{Definition }
\newdefinition{remark}[theorem]{Remark }
\newdefinition{example}[theorem]{Example }
\newproof{proof}{Proof}

\begin{document}

\begin{frontmatter}

\title{Bipolar fuzzy relation equations systems \\based on the product t-norm}

\author
{M. Eugenia Cornejo, David Lobo, Jes\'us Medina}

\address
{Department of Mathematics,
 University of  C\'adiz. Spain\\
\texttt{\{mariaeugenia.cornejo,david.lobo,jesus.medina\}@uca.es}}

\begin{abstract}

Bipolar fuzzy relation equations arise as a generalization of fuzzy relation equations considering unknown variables together with their logical connective negations. The occurrence of a variable and the occurrence of its negation simultaneously can give very useful information for certain frameworks where the human reasoning plays a key role. Hence, the resolution of bipolar fuzzy relation equations systems is a research topic of great interest.

This paper focuses on the study of bipolar fuzzy relation equations systems based on the max-product t-norm composition. Specifically, the solvability and the algebraic structure of the set of solutions of these bipolar equations systems will be studied, including the case in which such systems are composed of equations whose independent term be equal to zero. As a consequence, this paper complements the contribution carried out by the authors on the solvability of bipolar max-product fuzzy relation equations~\cite{CLM:JCAM2018}.
\end{abstract}

\begin{keyword}
Bipolar fuzzy relation equation, max-product t-norm composition, negation operator, fuzzy set.
\end{keyword}
\end{frontmatter}

\section{Introduction}
A broad development of the theory and applications of fuzzy relational equations (FREs), based on different max-t-norm compositions, has been carried out since they were introduced by Sanchez in the 1980s~\cite{sanchez76,sanchez79}. In the literature, we can find many papers dealing with the resolution of (systems of) FREs defined with either the max-min composition~\cite{chenwang07,chenwang02,Peeva:2004,Yeh2008}, the max-product composition~\cite{BOURKE1998,Loetamonphong:99,Markovskii,Peeva2007}, the max-Archimedean t-norm composition~\cite{LinJL,STAMOU2001,Wu2008} or other different compositions~\cite{Kohout80,Radim2012,baets99_eq,dm:mare,dm:ins2014,Medina2016,Ignjatovic20151,Medina2017:ija,Medina2017:ins402,Peeva2016,perfilieva08_fss}. In what regards to the applied perspective, it is important to emphasize that FREs have increased the range of applications of fuzzy sets theory. The compression and decompression of images and videos~\cite{Loia2005}, the modeling of fuzzy inference systems in fuzzy control~\cite{Ciaramella2006146} and the representation of restrictions in optimization problems~\cite{Li2008, Li2009,molai2010,yang2016}, among others, are some of the most recent applications of FREs based on max-t-norm compositions. FREs have also been related to other mathematical frameworks, such as fuzzy formal concept analysis~\cite{dm:mare,dm:ins2014}.

A new type of (system of) fuzzy relation equations arises when equations contain unknown variables together with their logical negations simultaneously. This new kind of equations are called bipolar fuzzy relation equations. Considering a negation operator provides the standard fuzzy relation equations with flexibility in the applications, as it is shown in~\cite{Freson2013, Hu2016,Li2014,Liu2015, Zhou2016}, where bipolar fuzzy relations equations have been successfully used in optimization problems.
To the best of our knowledge, the specific literature on the resolution of (systems of) bipolar fuzzy relation equations is really limited. A detailed analysis on the solvability of (systems) of bipolar max-min FREs with the standard negation is presented in~\cite{Li2016}. Following this research line, we can also find a wide study on the resolution of (systems) of bipolar max-product FREs with the product negation in~\cite{CLM:JCAM2018}. Clearly, the behaviour of the minimum t-norm and the product t-norm is different, therefore the results obtained in~\cite{CLM:JCAM2018} are markedly different from those presented in~\cite{Li2016}. 

In this paper, we are interested in continuing with the work done in~\cite{CLM:JCAM2018}. A characterization theorem for the solvability of (systems of) bipolar max-product FREs with the product negation and the properties related to the algebraic structure of the set of solutions have been introduced in~\cite{CLM:JCAM2018}. 
However, the solvability of (systems of) bipolar max-product FREs, with some independent term equal to zero, is an open problem to be solved.

Three different parts can be distinguished in our contribution. The first part introduces the notion of bipolar max-product fuzzy relation equation with the product negation. A characterization on the solvability of such equations is given and different properties associated with the existence of the greatest/least solution or a finite number of maximal/minimal solutions for these last equations are introduced. The second part shows under what conditions a bipolar max-product fuzzy relation equations system is solvable and presents the algebraic structure of the set of solutions of the solvable systems. {Both parts take into consideration the possibility of the independent term in the equations can be zero, which gives an added value to the work done by the authors in~\cite{CLM:JCAM2018}.} The third part includes a toy example which exposes a practical application. Finally, some conclusions and prospects for future work are included.

\section{Bipolar FREs based on the product t-norm}\label{sec:bmpn}

A detailed study on the resolution of bipolar max-product fuzzy relation equations with the product negation was carried out in~\cite{CLM:JCAM2018}, where  a characterization theorem for the solvability of such bipolar FREs was presented. The conditions to guarantee when a solvable bipolar max-product FRE has a greatest/least solution or a finite number of maximal/minimal solutions were also given in~\cite{CLM:JCAM2018}. However, the results related to the existence of the greatest/least solution or a finite number of maximal/minimal solutions do not consider bipolar max-product FREs whose independent term is zero. In this section, we will focus on this task.

Bipolar max-product FREs with the product negation are given from max-product FREs considering unknown variables together with their logical negations simultaneously, in that case the product negation. The formal definition of a bipolar max-product fuzzy relation equation with the product negation is introduced below. 

\begin{definition}\label{def:bmpp}
Let $a_j^+,a_j^-,b\in[0,1]$ and $x_j$ be an unknown variable belonging to $[0,1]$, for all $j\in\{1,\dots,m\}$, $\ast$   the product t-norm, $\vee$   the maximum operator and $n_P$   the product negation defined as $n_P(0)=1$ and $n_P(x)=0$, for all $x\in\,\,]0,1]$. Equation~\eqref{eq:bmpp} is called \emph{bipolar max-product fuzzy relation equation with the product negation}.
  \begin{equation}\label{eq:bmpp}
    \bigvee_{j=1}^m(a_j^+ \ast x_j) \vee (a_j^-\ast n_P (x_j))=b
  \end{equation}
\end{definition}

In the following, we will characterize the solvability of bipolar fuzzy relation equations defined previously, that is, we will provide the sufficient and necessary conditions under which bipolar max-product FREs with the product negation are solvable.

\begin{theorem}[\cite{CLM:JCAM2018}]\label{th:solvabilitybmpp}
Let $a_j^+,a_j^-,b\in[0,1]$ and $x_j$ be an unknown variable belonging to $[0,1]$, for all $j\in\{1,\dots,m\}$. The bipolar max-product fuzzy relation equation given by Equation~\eqref{eq:bmpp} is solvable if and only if one of the following statements is verified:
  \begin{itemize}
    \item[(a)] If $b=0$, then either $a_j^+=0$ or $a_j^-=0$, for each $j\in\{1,\dots,m\}$.
    \item[(b)] If $b\neq0$, then either $b\leq\max\{a_j^+\!\!\mid\! j\!\in\!\{1,\dots,m\}\}$ or there exists $k\in\{1,\dots,m\}$ such that $a_k^-=b$.
  \end{itemize}
\end{theorem}

Different cases have to be analyzed in order to ensure the existence of the greatest/least solution and the set of maximal/minimal solutions of a solvable bipolar max-product  FRE whose independent term is different from zero. It is convenient to remind the definition of the residuated implication associated with the product operator, that is $z\leftarrow_P x = \min(1,z/x)$, for all $x,z\in[0,1]$. It is also needed to mention that the operators $*$ and $\leftarrow_P$ satisfy the adjoint property, that is, $x*y\leq z$ if and only if $y\leq z\leftarrow_P x$, being $x,y,z\in[0,1]$. The residuated implication $\leftarrow_P$ and the adjoint property will play an important role throughout the paper.

\begin{theorem}[\cite{CLM:JCAM2018}]
Given $a_j^+,a_j^-\in[0,1]$, $b\in\, ]0,1]$, $x_j$ an unknown variable belonging to  $[0,1]$, for each $j\in\{1,\dots,m\}$, and a solvable bipolar max-product FRE as in Equation~\eqref{eq:bmpp}, then the following statements hold:
\begin{enumerate}
\item[(1)] If $b\leq\max\{a_j^+\mid j\in\{1,\dots,m\}\}$, then the set of solutions of Equation~\eqref{eq:bmpp} has a greatest element. The greatest solution is given by the tuple $(b\leftarrow_P a_1^+,\dots,b\leftarrow_P a_m^+)$.
\item[(2)] If $a_j^+< b$ for each $j\in\{1,\dots,m\}$, then the number of maximal solutions of Equation~\eqref{eq:bmpp} is finite. The set of maximal solutions of Equation~\eqref{eq:bmpp} is given by:
    \[\{(1,\dots,1, x_k,1,\dots,1)\mid  x_k=0 \hbox{ with } k\in K^-_P\}\]
where $K^-_P= \{k\in\{1,\dots,m\}\mid a_k^-=b$\}
\item[(3)] If there exists $k\in\{1,\dots,m\}$ such that $a_k^-=b$ and $a_j^-\leq b$, for each $j\in\{1,\dots,m\}$, then the set of solutions of Equation~\eqref{eq:bmpp} has a least solution. The least solution of Equation~\eqref{eq:bmpp} is $(0,\dots,0)$.
\item[(4)] If there exist $k_1,k_2\in\{1,\dots,m\}$ such that $a_{k_1}^-=b$ and $a_{k_2}^->b$, then the set of solutions of Equation~\eqref{eq:bmpp} has no minimal elements.
\item[(5)] If $a_j^-\neq b$ for each $j\in\{1,\dots,m\}$, then the number of minimal solutions of Equation~\eqref{eq:bmpp} is finite. The set of minimal solutions of Equation~\eqref{eq:bmpp} is given by:
    $$\{(0,\dots,0, x_k,0,\dots,0)\mid  x_k=b\leftarrow_P a_k^+ \hbox{ with } k\in K^+_P\}$$
    where  $K^+_P=\{k\in\{1,\dots,m\}\mid a_k^+\geq b\text{ and }a_j^- < b \text{ for each }j\neq k\}$.
\end{enumerate}
\end{theorem}

The following theorem shows that a solvable bipolar max-product FRE whose independent term is zero has always  {a} greatest solution. In addition, this result establishes the conditions under which such equation either  has   {a} least solution or it does not have minimal solutions.

\begin{theorem}
Given $a_j^+,a_j^-\in[0,1]$, $x_j$ an unknown variable belonging to $[0,1]$, for each $j\in\{1,\dots,m\}$, and a solvable bipolar max-product FRE given by:
  \begin{equation}\label{eq:bmppzero}
    \bigvee_{j=1}^m(a_j^+ \ast x_j) \vee (a_j^-\ast n_P (x_j))=0
  \end{equation}
then the following statements hold:
\begin{enumerate}

\item[(1)] The greatest solution of Equation~\eqref{eq:bmppzero} is given by the tuple $(\hat{x}_1,\dots,\hat{x}_m)$ which is defined as:
  \[\hat{x}_j= \left\{\begin{array}{lc}
             0 & \hbox{ if}\quad a_j^-=0 \\
             1 & \hbox{ if}\quad a_j^+=0
             \end{array}
  \right.\]
for each $j\in\{1,\dots,m\}$.
\item[(2)] The least solution of Equation~\eqref{eq:bmppzero} is $(0,\dots,0)$ if and only if $a_j^-=0$, for each $j\in\{1,\dots,m\}$.
\item[(3)] If there exists $k\in\{1,\dots,m\}$ such that $a_{k}^->0$, then the set of solutions of Equation~\eqref{eq:bmppzero} has no minimal element.
\end{enumerate}
\end{theorem}

\begin{proof}
Since Equation~\eqref{eq:bmppzero} is solvable, by Theorem~\ref{th:solvabilitybmpp}, we can ensure that either $a_j^+=0$ or $a_j^-=0$, for each $j\in\{1,\dots,m\}$. Taking into account this fact, we will prove Statments (1), (2) and (3).
\begin{enumerate}
\item[(1)] We will demonstrate that the tuple $(\hat{x}_1,\dots,\hat{x}_m)$ defined as:
  \[\hat{x}_j= \left\{\begin{array}{lc}
             0 & \hbox{ if}\quad a_j^-=0 \\
             1 & \hbox{ if}\quad a_j^+=0
             \end{array}
  \right.\]
for each $j\in\{1,\dots,m\}$, is a solution of Equation~\eqref{eq:bmppzero}. 

Given $k\in\{1,\dots,m\}$, if $a_k^+=0$, then $\hat{x}_k=1$ and thus we obtain the following equality:
  \[(a_k^+ \ast \hat{x}_k) \vee (a_k^-\ast n_P (\hat{x}_k))=(0 \ast 1)\vee (a_k^-\ast 0)=0\]

Otherwise, if $a_k^-=0$, then $\hat{x}_k=0$, which leads us to the equality:
  \[
  (a_k^+ \ast \hat{x}_k) \vee (a_k^-\ast n_P (\hat{x}_k))=(a_k^+\ast 0)\vee(0 \ast 1)=0
  \]
Therefore, the tuple $(\hat{x}_1,\dots,\hat{x}_m)$ verifies that: 
  \[
  \bigvee_{j=1}^m(a_j^+ \ast \hat{x}_j) \vee (a_j^-\ast n_P (\hat{x}_j))=0
  \]
  and, as a consequence, it is a solution of Equation~\eqref{eq:bmppzero}. 
  
In the following, we will prove that the tuple $(\hat{x}_1,\dots,\hat{x}_m)$ is the greatest solution of Equation~\eqref{eq:bmppzero} by reduction to the absurd. We will suppose that there exists a tuple $(x_{1},\dots,x_{m})$ being solution of Equation~\eqref{eq:bmppzero} such that $(x_{1},\dots,x_{m})\not\leq(\hat{x}_1,\dots,\hat{x}_m)$. Then, we can guarantee that there exists $k\in\{1,\dots,m\}$ such that $x_{k}>\hat{x}_k$. 
\begin{itemize}
\item If $a_k^+=0$, then by definition of the tuple $(\hat{x}_1,\dots,\hat{x}_m)$, we obtain that $x_{k}>\hat{x}_k=1$. This fact is a contradiction since, by hypothesis, the value $x_{k}$ belongs to $[0,1]$. 
\item Otherwise, if $a_k^-=0$ then by definition of the tuple $(\hat{x}_1,\dots,\hat{x}_m)$, we obtain that $x_{k}>\hat{x}_k=0$. Hence, we have that $(a_k^+ \ast x_{k}) \vee (a_k^-\ast n_P (x_{k}))=(a_k^+ \ast x_{k}) \vee (0\ast 0)=a_k^+ \ast x_{k}> 0$. Therefore
\[\bigvee_{j=1}^m(a_j^+ \ast \hat{x}_j) \vee (a_j^-\ast n_P (\hat{x}_j))\geq a_k^+ \ast x_{k}>0\]
This fact contradicts that $(x_{1},\dots,x_{m})$ be solution of Equation~\eqref{eq:bmppzero}.
\end{itemize}
As a consequence, we can ensure that $(\hat{x}_1,\dots,\hat{x}_m)$ is the greatest solution of Equation~\eqref{eq:bmppzero}.
 
\item[(2)] We will suppose that $a_j^-=0$, for each $j\in\{1,\dots,m\}$. Obviously, Equation~\eqref{eq:bmppzero} can be expressed in the following way:
\[\bigvee_{j=1}^m(a_j^+ \ast x_j) \vee (0\ast n_P(x_j))=\bigvee_{j=1}^m(a_j^+ \ast x_j)=0\]
Clearly, the tuple  $(0,\dots,0)$ is the only solution of Equation~\eqref{eq:bmppzero}. As a consequence, $(0,\dots,0)$ is the least solution of Equation~\eqref{eq:bmppzero}.  

In order to prove the counterpart, we will suppose that $(0,\dots,0)$ is the least solution of Equation~\eqref{eq:bmppzero}. Therefore, we obtain that:
\[\bigvee_{j=1}^m(a_j^+ \ast 0) \vee (a_j^-\ast n_P(0))=\bigvee_{j=1}^m(a_j^-\ast n_P(0))=\bigvee_{j=1}^m a_j^-=0\]
Consequently, we can ensure that $a_j^-=0$, for each $j\in\{1,\dots,m\}$.

\item[(3)] We will suppose that there exists $k\in\{1,\dots,m\}$ such that $a_{k}^->0$ in Equation~\eqref{eq:bmppzero}, by Theorem~\ref{th:solvabilitybmpp}, $a_{k}^+=0$.  If $(x_{1},\dots,x_{k-1},x_{k}, x_{k+1},\dots,x_{m})$ is solution of Equation~\eqref{eq:bmppzero} then $x_k>0$. Clearly, the tuple given by $(x_{1},\dots,x_{k-1},\frac{x_{k}}{2}, x_{k+1},\dots,x_{m})$ is also solution of  Equation~\eqref{eq:bmppzero} verifying $(x_{1},\dots,x_{k-1},\frac{x_{k}}{2}, x_{k+1},\dots,x_{m})<(x_{1},\dots,x_{k-1},x_{k}, x_{k+1},\dots,x_{m})$. Since this procedure can be repeated indefinitely we can ensure that Equation~\eqref{eq:bmppzero} does not have minimal solutions.
\end{enumerate}
\end{proof}

In view of the result obtained in the previous theorem, we can ensure that the existence of the least solution can only be  guaranteed when the considered solvable bipolar max-product FRE as Equation~\eqref{eq:bmppzero} is actually a solvable max-product FRE with independent term equal to zero. 

After characterizing the solvability of bipolar max-product FREs with the product negation and studying the algebraic structure of the set of solutions, we are interested in the resolution of systems of bipolar max-product FREs with the product negation.

\section{Solving bipolar FREs systems from the product t-norm}\label{sec:bmpnsystems}

An initial study on the solvability of systems of bipolar max-product FREs with the product negation was presented in~\cite{CLM:JCAM2018}. Now, we will deepen in this study including results related to the algebraic structure of the complete set of solutions corresponding to an arbitrary solvable system of bipolar max-product FREs with the product negation. 

It is worth highlighting that the conditions required to guarantee the solvability of systems of bipolar max-product FREs  are significantly different from the obtained ones for only one  bipolar max-product FRE. Specifically, the resolution of an arbitrary system of bipolar max-product FREs is characterized by the existence of two index sets satisfying certain properties. The following definition includes the notion of feasible pair of index sets, which will play an important role from now on.

\begin{definition}\label{def:fp}
Let $m,n\in\mathbb{N}$, $a_{ij}^+,a_{ij}^-,b_i,\,x_j\in[0,1]$, for each $i\in\{1,\dots,n\}$ and $j\in\{1,\dots,m\}$. Consider the bipolar max-product  FREs system given by System~\eqref{sys:general_mxn}:
\begin{equation}\label{sys:general_mxn}
\bigvee_{j=1}^m(a_{ij}^+ \ast x_j) \vee (a_{ij}^-\ast n_P (x_j))=b_i \qquad i\in\{1,\dots,n\}
\end{equation}
A pair of index sets $(J^+,J^-)$ with $J^+,J^-\subseteq\{1,\dots,m\}$ is said to be \emph{feasible with respect to System~\eqref{sys:general_mxn}} if\footnote{Notice that, by definition, $J^+\uplus J^-=\{1,\dots,m\}$ if and only if $J^+\cup J^-=\{1,\dots,m\}$ and $J^+\cap J^-=\varnothing$.} $J^+\uplus J^-=\{1,\dots,m\}$ and for each $i\in\{1,\dots,n\}$:
  \begin{enumerate}[label=(\alph*)]
  \item If $b_i=0$, then $a_{ij}^+=0$ for each $j\in J^+$ and $a_{ij}^-=0$ for each $j\in J^-$.
  \item If $b_i>0$, then one of the following statements is verified:
    \begin{enumerate}
      \item[(b1)] there exists $j\in J^+$ such that $a_{ij}^+\geq b_i$ and $b_i\leftarrow_P a_{ij}^+\leq b_h\leftarrow_P a_{hj}^+$, for each $h\in\{1,\dots,n\}$.
      \item[(b2)] there exists $j\in J^-$ such that $a_{ij}^-=b_i$ and the inequality $a_{hj}^-\leq b_h$ is satisfied, for each $h\in\{1,\dots,n\}$.
    \end{enumerate}
  \end{enumerate}
\end{definition}

As we mentioned above, the solvability of bipolar max-product FREs systems with the product negation is characterized by the existence of feasible pairs of index sets, as the following theorem shows.

\begin{theorem}\label{th:sys_general_mxn}
Let $a_{ij}^+,a_{ij}^-,b_i,\,x_j\in[0,1]$, for each $i\in\{1,\dots,n\}$ and $j\in\{1,\dots,m\}$. The bipolar max-product  FRE system given by System~\eqref{sys:general_mxn} is solvable if and only if there exists at least a feasible pair with respect to System~\eqref{sys:general_mxn}.
\end{theorem}
\begin{proof}
Suppose that there exists a feasible pair $(J^+, J^-)$ with respect to System~\eqref{sys:general_mxn} and we will prove that the tuple $(x_1,\dots,x_m)$ defined, for each $j\in\{1,\dots,m\}$, as:
  \[x_j= \left\{\begin{array}{lc}
             0 & \hbox{ if}\quad j\in J^- \\
             \min\{b_h\leftarrow_P a_{hj}^+\mid h\in\{1,\dots,n\}\} & \hbox{ if}\quad j\in J^+
             \end{array}
  \right.\]
is solution of System~\eqref{sys:general_mxn}. Fixed $i\!\in\!\{1,\dots,n\}$, we will distinguish two cases:
\begin{itemize}
\item Case $b_i=0$: According to Definition~\ref{def:fp}, we obtain that $a_{ij}^+=0$ for each $j\in J^+$ and $a_{ij}^-=0$ for each $j\in J^-$. By the definition of $x_j$, we have that $x_j=0$, for each $j\in J^-$. Hence, for each $j\in J^-$, the next chain of equalities is satisfied:
\[(a_{ij}^+ \ast x_j) \vee (a_{ij}^-\ast n_P (x_j))=(a_{ij}^+ \ast 0) \vee (0*1)=0\]
Considering again the definition of $x_j$, for each $j\in J^+$, we can ensure that $x_j=\min\{b_h\leftarrow_P a_{hj}^+\mid h\in\{1,\dots,n\}\}$, and therefore $x_j>0$. Thus,  for each $j\in J^+$, the following chain of equalities is verified:
\[(a_{ij}^+ \ast x_j) \vee (a_{ij}^-\ast n_P (x_j))=(0 \ast x_j) \vee (a_{ij}^-\ast0)=0\]
Due to $J^+\uplus J^-=\{1,\dots,m\}$, we conclude that
\[\bigvee_{j=1}^m(a_{ij}^+ \ast x_j) \vee (a_{ij}^-\ast n_P (x_j))=0=b_i\]
That is, the tuple  $(x_1,\dots,x_m)$ satisfies the $i$-th equation of System~\eqref{sys:general_mxn}.

\item Case $b_i>0$: The proof of the tuple  $(x_1,\dots,x_m)$ satisfies the $i$-th equation of System~\eqref{sys:general_mxn} can be found in Theorem 20 of~\cite{CLM:JCAM2018}.
\end{itemize}
Therefore, the $i$-th equation of System~\eqref{sys:general_mxn} given by:
\[\bigvee_{j=1}^m(a_{ij}^+ \ast x_{j}) \vee (a_{ij}^-\ast n_P (x_{j}))=b_i\]
is satisfied. Following an analogous reasoning for each $i\in\{1,\dots,n\}$, we can conclude that $(x_{1},\dots,x_{m})$ is a solution of System~\eqref{sys:general_mxn}.

In order to prove the counterpart, suppose that System~\eqref{sys:general_mxn} is solvable and we will demonstrate that there exists at least a feasible pair with respect to System~\eqref{sys:general_mxn}. Given a solution $(x_{1},\dots,x_{m})$ of System~\eqref{sys:general_mxn}, let us define two index sets $J^+$ and $J^-$ as follows:
\begin{eqnarray*}
J^+&=&\{j\in\{1,\dots,m\}\mid x_j>0\}\\
J^-&=&\{j\in\{1,\dots,m\}\mid x_j=0\}
\end{eqnarray*}
Clearly, $J^+\uplus J^-=\{1,\dots,m\}$. That is, the index sets $J^+$ and $J^-$ sastify that $J^+\cap J^-=\varnothing$ and  $J^+\cup J^-=\{1,\dots,m\}$. Fixed $i\in\{1,\dots,n\}$, we will distinguish two cases:
\begin{itemize}
\item Case $b_i=0$: We obtain that $(a_{ij}^+ \ast x_{j}) \vee (a_{ij}^-\ast n_P (x_{j}))=0$, for each $j\in\{1,\dots,m\}$. By the definition of the maximum operator, we have that $a_{ij}^+ \ast x_{j}=0$ and  $a_{ij}^-\ast n_P (x_{j})=0$, for each $j\in\{1,\dots,m\}$. On the one hand, for each $j\in J^+$, we have that $x_j>0$ and as a consequence, we deduce that $a_{ij}^+$ has to be equal to zero in order to the equality $a_{ij}^+ \ast x_{j}=0$ be satisfied. On the other hand, for each $j\in J^-$, we have that $x_j=0$ and therefore, we obtain that $a_{ij}^-$ has to be equal to zero in order to the equality $a_{ij}^-\ast n_P (x_{j})=0$ be satisfied. Consequently, $a_{ij}^+=0$ for each $j\in J^+$ and $a_{ij}^-=0$ $j\in J^-$.
\item Case $b_i>0$: The proof of the index sets $J^+$ and $J^-$ satisfy Statement (b) of Definition~\ref{def:fp} can be found in Theorem 20 of~\cite{CLM:JCAM2018}. 
\end{itemize}
By using an analogous reasoning for each $i\in\{1,\dots,n\}$, we can conclude that $(J^+, J^-)$ is a feasible pair with respect to System~\eqref{sys:general_mxn}.
\end{proof}

It is important to mention that the previous theorem provides an added value to Theorem 20 presented by the authors in~\cite{CLM:JCAM2018}, since the characterization of the solvability of bipolar max-product FREs systems is completed with the consideration of systems whose equations can take the value zero in the independent term.

The following step in our study will be to know when a bipolar max-product FRE system with the product negation has either a greatest/least solution or a finite number of maximal/minimal solutions. Specifically, in what regards the existence of the greatest solution or maximal solution, the next result shows that the algebraic structure of a bipolar max-product FRE system is closely related to the algebraic structure of the set of feasible pairs with respect to that system.

\begin{theorem}\label{th:sys_maximales}
  Let $a_{ij}^+,a_{ij}^-,b_i,\,x_j\in[0,1]$, for each $i\in\{1,\dots,n\}$ and $j\in\{1,\dots,m\}$. Consider that System~\eqref{sys:general_mxn}
  is a solvable bipolar max-product  FRE system, and let $S$, $S^+$ be the sets defined as:
\[S=\{(J^+,J^-)\mid J^+,J^-\subseteq\{1,\dots,m\}, (J^+,J^-) \hbox{ is a feasible pair w.r.t System~\eqref{sys:general_mxn}}\}\]
  \[S^+=\{J^+\mid(J^+,J^-)\in S\}\]
  Then, the following statements hold:
  \begin{enumerate}
  \item[(1)] If $S^+$ has a greatest element, then System~\eqref{sys:general_mxn} has a greatest solution.
  \item[(2)]  The number of maximal solutions of System~\eqref{sys:general_mxn} coincides with the number of maximal elements of $S^+$.
  \end{enumerate}
\end{theorem}
\begin{proof}
First of all, we will prove Statement (1). Suppose that $S^+$ has a greatest element, that is, there exists $\hat{J}^+\in S^+$ such that $J^+\subseteq\hat{J}^+$ for each $J^+\in S^+$, and consider the tuple $(\hat{x}_1,\dots,\hat{x}_m)$ defined, for each $j\in\{1,\dots,m\}$, as follows:
\[\hat{x}_j= \left\{\begin{array}{lc}
             0 & \hbox{ if}\quad j\in \hat{J}^- \\
             \min\{b_h\leftarrow_P a_{hj}^+\mid h\in\{1,\dots,n\}\} & \hbox{ if}\quad j\in \hat{J}^+
             \end{array}
  \right.\]
being $\hat{J}^-=\{1,\dots,m\}\setminus \hat{J}^+$. We will see by reduction to the absurd that $(\hat{x}_1,\dots,\hat{x}_m)$ is the greatest solution of System~\eqref{sys:general_mxn}. 

Suppose that there exists a solution $(x_1,\dots,x_m)$ of System~\eqref{sys:general_mxn} such that $(x_1,\dots,x_m)\not\leq (\hat{x}_1,\dots,\hat{x}_m)$. That is, there exists $k\in\{1,\dots,m\}$ such that $x_k>\hat{x}_k$. Now, let us distinguish the following two cases and we will obtain a contradiction from both of them:
\begin{itemize}
\item Case $k\in \hat{J}^+$: By definition of the tuple $(\hat{x}_1,\dots,\hat{x}_m)$, we obtain that $\hat{x}_k=\min\{b_h\leftarrow_P a_{hk}^+\mid h\in\{1,\dots,n\}\}$. Consider fixed the index $i\in\{1,\dots,n\}$ such that $b_i\leftarrow_P a_{ik}^+=\min\{b_h\leftarrow_P a_{hk}^+\mid h\in\{1,\dots,n\}\}$. Since $x_k>\hat{x}_k=b_i\leftarrow_P a_{ik}^+$, by the adjoint property, we can ensure that $a_{ik}^+*x_k> b_i$. Therefore, the following chain of inequalities holds:
\[\bigvee_{j=1}^m(a_{ij}^+ \ast x_j) \vee (a_{ij}^-\ast n_P (x_j))\geq a_{ik}^+*x_k> b_i\]
Consequently, $i$-th equation is not satisfied and thus $(x_1,\dots,x_m)$ is not a solution of System~\eqref{sys:general_mxn}, which is a contradiction.
\item Case $k\in \hat{J}^-$: By definition of the tuple $(\hat{x}_1,\dots,\hat{x}_m)$, we have that $\hat{x}_k=0$. In addition, $x_k>0$ since $x_k>\hat{x}_k$ by hypothesis. Since $(x_1,\dots,x_m)$ is a solution of System~\eqref{sys:general_mxn}, applying Theorem~\ref{th:sys_general_mxn}, we can find two index sets $J^+, J^-$ defined as follows:
\begin{eqnarray*}
J^+&=&\{j\in\{1,\dots,m\}\mid x_j>0\}\\
J^-&=&\{j\in\{1,\dots,m\}\mid x_j=0\}
\end{eqnarray*}
such that $(J^+,J^-)\in S$. Due to the fact that $k\in J^+$ and $\hat{J}^+\cap \hat{J}^-=\varnothing$, we can assert that $J^+\not\subseteq \hat{J}^+$, in contradiction with the hypothesis.
\end{itemize}
Hence, the tuple $(\hat{x}_1,\dots,\hat{x}_m)$ is the greatest solution of System~\eqref{sys:general_mxn}.

Now we will prove  Statement (2), that is, we will see that the number of maximal solutions of System~\eqref{sys:general_mxn} coincides with the number of maximal elements of $S^+$. To reach this conclusion, we will denote the set of maximal elements of $S^+$ as $A=\{\hat{J}^+\in S^+\mid \hat{J}^+\not\subset J^+\hbox{ for each }J^+\in S^+\}$, the set of maximal solutions of System~\eqref{sys:general_mxn} as $B$ and we will see that the mapping $f\colon A\to B$, which associates each $\hat{J}^+\in A$ with the tuple $(\hat{x}_1,\dots,\hat{x}_m)$ defined as:
\[\hat{x}_j= \left\{\begin{array}{lc}
             0 & \hbox{ if}\quad j\notin \hat{J}^+ \\
             \min\{b_h\leftarrow_P a_{hj}^+\mid h\in\{1,\dots,n\}\} & \hbox{ if}\quad j\in \hat{J}^+
             \end{array}
  \right.\]
for each $j\in\{1,\dots,m\}$, is a bijection. 

First of all, we will see that $f$ is well-defined. Given $\hat{J}^+\in A$, we will prove that $f(\hat{J}^+)=(\hat{x}_1,\dots,\hat{x}_m)$ belongs to $B$, that is, that $f(\hat{J}^+)$ is a maximal solution of System~\eqref{sys:general_mxn}. Applying Theorem~\ref{th:sys_general_mxn}, we can ensure that $(\hat{x}_1,\dots,\hat{x}_m)$ is a solution of System~\eqref{sys:general_mxn} since $(\hat{J}^+,\hat{J}^-)\in S$. It remains to prove that $(\hat{x}_1,\dots,\hat{x}_m)$ is a maximal solution. We will reach to this conclusion by reduction to the absurd. Consequently, suppose that there exists a solution $(x_1,\dots,x_m)$ of System~\eqref{sys:general_mxn} such that $(x_1,\dots,x_m)> (\hat{x}_1,\dots,\hat{x}_m)$. Clearly, we can assert that $x_j\geq\hat{x}_j$, for each $j\in\{1,\dots,m\}$, and that there exists $k\in\{1,\dots,m\}$ such that $x_{k}>\hat{x}_{k}$. 
Suppose that $k\in \hat{J}^+$ and consider fixed $i\in\{1,\dots,n\}$ such that $b_i\leftarrow_P a_{ik}^+=\min\{b_h\leftarrow_P a_{hk}^+\mid h\in\{1,\dots,n\}\}$, we obtain that $\hat{x}_{k}=b_i\leftarrow_P a_{ik}^+$. From the adjoint property 
and taking into account that  $x_{k}>\hat{x}_{k}$, we can assert that $x_{k}>b_i\leftarrow_P a_{ik}^+$ implies that $a_{ik}^+*x_{k}> b_i$. Hence, the chain of inequalities below holds:
\[\bigvee_{j=1}^m(a_{ij}^+ \ast x_j) \vee (a_{ij}^-\ast n_P (x_j))\geq a_{ik}^+*x_k> b_i\]
That is, $i$-th equation is not satisfied, and therefore $(x_1,\dots,x_m)$ is not a solution of System~\eqref{sys:general_mxn}, in contradiction with the hypothesis.
Therefore, we can assert that $k\notin \hat{J}^+$. Hence, $\hat{x}_{k}=0$ by definition and $x_{k}>\hat{x}_{k}=0$.

Now, we consider two index sets $J^+$ and $J^-$ defined in the following way:
\begin{eqnarray*}
J^+&=&\{j\in\{1,\dots,m\}\mid x_j>0\}\\
J^-&=&\{j\in\{1,\dots,m\}\mid x_j=0\}
\end{eqnarray*}

Following an analogous reasoning to the proof in Theorem~\ref{th:sys_general_mxn}, since $(x_1,\dots,x_m)$ is a solution of System~\eqref{sys:general_mxn}, we can assert that $(J^+,J^-)\in S$. In particular, we have that $J^+\in S^+$.
On the one hand, we obtain that $\hat{J}^+\subseteq J^+$ since $x_j\geq\hat{x}_j$, for each $j\in\{1,\dots,m\}$. On the other hand, we obtain that $k\in J^+$, since $x_{k}>0$. Taking into account that $k\in \hat{J}^-$ and $\hat{J}^+\uplus\hat{J}^-=\{1,\dots,m\}$, we have that $k\notin \hat{J}^+$. Hence, we deduce that $\hat{J}^+\subset J^+$, which is a contradiction because the index set $\hat{J}^+$ was supposed to be a maximal element of the set $S^+$.

Therefore, we conclude that $(\hat{x}_1,\dots,\hat{x}_m)$ is a maximal solution of System~\eqref{sys:general_mxn}, that is, $f(\hat{J}^+)=(\hat{x}_1,\dots,\hat{x}_m)$ belongs to $B$. Thus, the mapping $f$ is well-defined.

Hereinafter, we will see that $f$ is a bijection between $A$ and $B$. Let us prove that $f$ is order-embedding. Consider $\hat{J}^+_1,\hat{J}^+_2\in A$ with $\hat{J}^+_1\neq \hat{J}^+_2$, being $f(\hat{J}^+_1)=(\hat{x}_1^1,\dots,\hat{x}_m^1)$ and $f(\hat{J}^+_2)=(\hat{x}_1^2,\dots,\hat{x}_m^2)$. Without lost of generality, we can ensure that there exists $k\in\{1,\dots,m\}$ such that $k\in \hat{J}^+_1$ and $k\notin \hat{J}^+_2$, and then $\hat{x}_k^1>0$ while $\hat{x}_k^2=0$. 
This fact implies that $(\hat{x}_1^1,\dots,\hat{x}_m^1)\neq(\hat{x}_1^2,\dots,\hat{x}_m^2)$. 

Lastly, we demonstrate that $f$ is onto. Given $(\hat{x}_1,\dots,\hat{x}_m)\in B$, that is, a maximal solution of System~\eqref{sys:general_mxn}, we will obtain a set $\hat{J}^+\in A$ such that $f(\hat{J}^+)=(\hat{x}_1,\dots,\hat{x}_m)$. Consider the sets $\hat{J}^+, \hat{J}^-$ defined as follows:
\begin{eqnarray*}
\hat{J}^+&=&\{j\in\{1,\dots,m\}\mid \hat{x}_j>0\}\\
\hat{J}^-&=&\{j\in\{1,\dots,m\}\mid \hat{x}_j=0\}
\end{eqnarray*}
Since $(\hat{x}_1,\dots,\hat{x}_m)$ is a solution of System~\eqref{sys:general_mxn}, applying Theorem~\ref{th:sys_general_mxn}, we deduce that $(\hat{J}^+,\hat{J}^-)\in S$ and $\hat{J}^+\in S^+$. In order to prove that $\hat{J}^+\in A$, we will proceed by reduction to the absurd. 

Suppose that there exists $J^+\in S^+$ such that $\hat{J}^+\subset J^+$. Then, we define the tuple $(x_1,\dots,x_m)$ in the following form:
\[x_j= \left\{\begin{array}{lc}
             0 & \hbox{ if}\quad j\notin J^+ \\
             \min\{b_h\leftarrow_P a_{hj}^+\mid h\in\{1,\dots,n\}\} & \hbox{ if}\quad j\in J^+
             \end{array}
  \right.\]
We obtain that $(J^+,J^-)\in S$, being $J^-=\{1,\dots,m\}\setminus J^+$, since $J^+\in S^+$. Applying Theorem~\ref{th:sys_general_mxn}, we obtain that $(x_1,\dots,x_m)$ is a solution of System~\eqref{sys:general_mxn}. Now, taking into account the definition of the tuples $(x_1,\dots,x_m)$ and $(\hat{x}_1,\dots,\hat{x}_m)$, we can observe that: 

\begin{itemize}
\item For each $j\in\hat{J}^-\cap J^-$, by definition $\hat{x}_j=x_j=0$. 
\item Given $j\in J^+$, we can deduce that $\hat{x}_j\leq x_j$. In fact, if $\hat{x}_j> x_j$, then there exists $i\in\{1,\dots,n\}$ such that $\hat{x}_j>b_i\leftarrow_P a_{ij}^+$. According to the adjoint property, we obtain that $a_{ij}^+*\hat{x}_j>b_i$. From this fact, we conclude that the $i$-th equation in System~\eqref{sys:general_mxn} is not satisfied. Thus, the tuple $(\hat{x}_1,\dots,\hat{x}_m)$ is not a solution of System~\eqref{sys:general_mxn}, in contradiction with the hypothesis. Therefore we can assert that $\hat{x}_j\leq x_j$ for each $j\in J^+$. Moreover, since $\hat{J}^+\subset J^+$, we obtain that $\hat{x}_j\leq x_j$ for each $j\in \hat{J}^+$.
\item Taking into account $\hat{J}^+\subset J^+$, we can ensure that there exists $k\in J^+$ such that $k\notin\hat{J}^+$. According to the definition of $x_k$ and $\hat{x}_k$, we obtain that $x_k>\hat{x}_k=0$.
\end{itemize}
Hence, we can assert that $(x_1,\dots,x_m)>(\hat{x}_1,\dots,\hat{x}_m)$, which contradicts the fact that $(\hat{x}_1,\dots,\hat{x}_m)$ is a maximal solution of System~\eqref{sys:general_mxn}. Consequently, we conclude that there is no $J^+\in S^+$ such that $\hat{J}^+\subset J^+$, or equivalently, $\hat{J}^+$ is a maximal element in $S^+$, that is $\hat{J}^+\in A$.

To finish with this demonstration, we will see that $f(\hat{J}^+)=(\hat{x}_1,\dots,\hat{x}_m)$. Firstly, notice that $\hat{x}_j$ is straightforwardly equal to $0$, for each $j\notin \hat{J}^+$. Now, suppose that there exists $k\in \hat{J}^+$ such that $\hat{x}_k\neq \min\{b_h\leftarrow_P a_{hk}^+\mid h\in\{1,\dots,n\}\}$. Given $i\in\{1,\dots,n\}$ such that $b_i\leftarrow_P a_{ik}^+=\min\{b_h\leftarrow_P a_{hk}^+\mid h\in\{1,\dots,n\}\}$, then one and only one of the next two cases is hold:
\begin{itemize}
\item Case $\hat{x}_k>b_i\leftarrow_P a_{ik}^+$: By the adjoint property, 
we can assert that $\hat{x}_k>b_i\leftarrow_P a_{ik}^+$ implies $a_{ik}^+*\hat{x}_k> b_i$. As a result, the chain of inequalities below holds:
\[\bigvee_{j=1}^m(a_{ij}^+ \ast \hat{x}_j) \vee (a_{ij}^-\ast n_P (\hat{x}_j)\geq a_{ik}^+*\hat{x}_k> b_i\]
Thus, the tuple $(\hat{x}_1,\dots,\hat{x}_m)$ is not a solution of System~\eqref{sys:general_mxn}, in contradiction with the hypothesis.
\item Case $\hat{x}_k<b_i\leftarrow_P a_{ik}^+$: We have that  a solution of System~\eqref{sys:general_mxn} is $(\hat{x}_1,\dots,\hat{x}_{k-1},b_i\leftarrow_P a_{ik}^+,\hat{x}_{k+1},\dots,\hat{x}_m)$, which is clearly strictly greater than $(\hat{x}_1,\dots,\hat{x}_m)$. This fact contradicts that $(\hat{x}_1,\dots,\hat{x}_m)$ is a maximal solution of System~\eqref{sys:general_mxn}.
\end{itemize}
Hence, we can ensure that $\hat{x}_j= \min\{b_h\leftarrow_P a_{hj}^+\mid h\in\{1,\dots,n\}\}$, for each $j\in \hat{J}^+$. To sum up, we have shown that the tuple $(\hat{x}_1,\dots,\hat{x}_m)$ is given, for each $j\in\{1,\dots,m\}$, by
\[\hat{x}_j= \left\{\begin{array}{lc}
             0 & \hbox{ if}\quad j\notin \hat{J}^+ \\
             \min\{b_h\leftarrow_P a_{hj}^+\mid h\in\{1,\dots,n\}\} & \hbox{ if}\quad j\in \hat{J}^+
             \end{array}
  \right.\]
That is, $f(\hat{J}^+)=(\hat{x}_1,\dots,\hat{x}_m)$. Consequently, we conclude that $f$ is a bijection between $A$ and $B$, and thus, the number of maximal solutions of System~\eqref{sys:general_mxn} coincides with the number of maximal elements of $S^+$.
\end{proof}

The following corollary arises as a consequence of Theorem~\ref{th:sys_maximales}.

\begin{corollary}\label{cor:sys_maximales}
  Let $a_{ij}^+,a_{ij}^-,b_i,\,x_j\in[0,1]$, for each $i\in\{1,\dots,n\}$ and $j\in\{1,\dots,m\}$. Consider that System~\eqref{sys:general_mxn}
  is a solvable bipolar max-product  FRE system, and let $S$, $S^+$ be the sets defined as:
   \[S=\{(J^+,J^-)\mid J^+,J^-\subseteq\{1,\dots,m\}, (J^+,J^-) \hbox{ is a feasible pair w.r.t System~\eqref{sys:general_mxn}}\}\]
  \[S^+=\{J^+\mid(J^+,J^-)\in S\}\]
  Consider the mapping $f\colon S^+\to [0,1]^m$ which associates each $J^+\in S^+$ with the tuple $(x_1,\dots,x_m)$ defined, for each $j\in\{1,\dots,m\}$, by
\[x_j= \left\{\begin{array}{lc}
             0 & \hbox{ if}\quad j\notin J^+ \\
             \min\{b_h\leftarrow_P a_{hj}^+\mid h\in\{1,\dots,n\}\} & \hbox{ if}\quad j\in J^+
             \end{array}
  \right.\]
  Then, the following statements hold:
  \begin{enumerate}
  \item[(1)] If $S^+$ has a greatest element $J^+$, then $f(J^+)$ is the greatest solution of System~\eqref{sys:general_mxn}.
  \item[(2)] Let 
  $M^+$ be the set of maximal elements of $S^+$. Then, the set of maximal solutions of System~\eqref{sys:general_mxn} is given by:
  \[\{f(J^+)\mid J^+\in M^+\}\]
  \end{enumerate}
\end{corollary}

The existence of the least solution and the set of minimal solutions of a solvable bipolar max-product FREs system is also studied distinguishing cases, as it is shown below.

\begin{theorem}\label{th:sys_minimales}
  Let $a_{ij}^+,a_{ij}^-,b_i,\,x_j\in[0,1]$, for each $i\in\{1,\dots,n\}$ and $j\in\{1,\dots,m\}$. Consider that System~\eqref{sys:general_mxn}
  is a solvable bipolar max-product FREs system and let $S$, $S^-$ be the sets defined as:
   \[S=\{(J^+,J^-)\mid J^+,J^-\subseteq\{1,\dots,m\}, (J^+,J^-) \hbox{ is a feasible pair w.r.t System~\eqref{sys:general_mxn}}\}\]
  \[S^-=\{J^-\mid(J^+,J^-)\in S\}\]
  Considering, for each $J^-\in S^-$, the set $I_{J^-}$ defined as:
  \[I_{J^-}=\{i\in\{1,\dots,n\}\mid b_i=0\hbox{ or }a_{ij}^-<b_i \hbox{ for each } j\in J^-\}\]
  we obtain the following statements:
  \begin{enumerate}
  \item If $\{1,\dots,m\}\in S^-$, then System~\eqref{sys:general_mxn} has a least solution.
  \item If $\{1,\dots,m\}\notin S^-$, the number of minimal solutions of System~\eqref{sys:general_mxn} coincides with the number of maximal elements  of the set $S^-$, 
   such that the following system has a unique solution:
  \begin{equation}\label{eq:sys_minimales}
	\bigvee_{\substack{j\in \{1,\dots,m\}\\ j\notin J^-}}(a_{ij}^+ \ast x_j) \vee (a_{ij}^-\ast n_P (x_j))=b_i,\qquad i\in I_{J^-}
  \end{equation}
  \end{enumerate}
\end{theorem}
\begin{proof}
Notice that if $\{1,\dots,m\}\in S^-$, then following an analogous reasoning to the proof given in Theorem~\ref{th:sys_general_mxn}, we obtain that the tuple $(0,\dots,0)$ is a solution of System~\eqref{sys:general_mxn} and clearly it is the least solution of System~\eqref{sys:general_mxn}. Hence, Statement (1) is straightforwardly satisfied.

In order to prove Statement (2), we will define $A$ as the set of maximal elements $J^-$ of $S^-$ such that System~\eqref{eq:sys_minimales} has only one solution and $B$ as the set of minimal solutions of System~\eqref{sys:general_mxn}. We have to see that the cardinal of $A$ coincides with the cardinal of $B$. To reach this target, we will define a mapping from $A$ to $B$ and we will see that it forms a bijection. Consider the mapping $f\colon A\to B$ such that each $\hat{J}^-\in A$ is associated with the tuple $(\hat{x}_1,\dots,\hat{x}_m)$ defined, for each $j\in\{1,\dots,m\}$, as
\[\hat{x}_j= \left\{\begin{array}{lc}
             0 & \hbox{ if}\quad j\in \hat{J}^-\\
             \min\{b_h\leftarrow_P a_{hj}^+\mid h\in\{1,\dots,n\}\} & \hbox{ if}\quad j\notin \hat{J}^- 
             \end{array}
  \right.\]
To begin with, we will see that $f$ is well defined. That is, given $\hat{J}^-\in A$, we will conclude that the tuple $f(\hat{J}^-)=(\hat{x}_1,\dots,\hat{x}_m)$ is a minimal solution of System~\eqref{sys:general_mxn}, and thus $(\hat{x}_1,\dots,\hat{x}_m)$ belongs to the set $B$. Clearly, if $\{1,\dots,m\}\notin S^-$ then $\hat{J}^+=\{1,\dots,m\}\setminus\hat{J}^-$, and following an analogous reasoning to the proof given in Theorem~\ref{th:sys_general_mxn}, we obtain that $(\hat{x}_1,\dots,\hat{x}_m)$ is a solution of System~\eqref{sys:general_mxn}. In order to see that it is a minimal solution, we are going to proceed by reduction to the absurd.

Suppose that there exists a solution $(x_1,\dots,x_m)$ of System~\eqref{sys:general_mxn} such that $(x_1,\dots,x_m)<(\hat{x}_1,\dots,\hat{x}_m)$. Let us consider that the set $\hat{J}^+$ has $l$ elements with $l\in\mathbb{N}$, and we denote this set as $\hat{J}^+=\{j_1,\dots,j_l\}$. Then, we are going to deduce that the corresponding tuples $(x_{j_1},\dots,x_{j_l})$ and $(\hat{x}_{j_1},\dots,\hat{x}_{j_l})$ are different and they both are solutions of System~\eqref{eq:sys_minimales}, in contradiction with the hypothesis.

According to $(x_1,\dots,x_m)<(\hat{x}_1,\dots,\hat{x}_m)$, clearly $(x_{j_1},\dots,x_{j_l})\leq(\hat{x}_{j_1},\dots,\hat{x}_{j_l})$. Furthermore, the inequality $(x_1,\dots,x_m)<(\hat{x}_1,\dots,\hat{x}_m)$ implies that there exists $k\in\{1,\dots,m\}$ such that $x_{k}<\hat{x}_{k}$. Notice that, $\hat{x}_{j}=0$ for each $j\in\hat{J}^-$ and the chain $0\leq x_{k}<\hat{x}_{k}$ holds, thus we obtain that $k\notin\hat{J}^-$. Therefore, by definition, $k\in\hat{J}^+$. As a consequence, we can assert that $(x_{j_1},\dots,x_{j_l})<(\hat{x}_{j_1},\dots,\hat{x}_{j_l})$, and therefore they are different tuples. It remains to demonstrate that they both are solutions of System~\eqref{eq:sys_minimales}. Clearly, if $I_{\hat{J}^-}=\varnothing$, then $(x_{j_1},\dots,x_{j_l})$ and $(\hat{x}_{j_1},\dots,\hat{x}_{j_l})$ are straightforwardly solutions of System~\eqref{eq:sys_minimales}, since there are no equations to be satisfied. Therefore, we can suppose from now on that $I_{\hat{J}^-}\neq\varnothing$. 

Notice that, as $(\hat{x}_1,\dots,\hat{x}_m)$ is a solution of System~\eqref{sys:general_mxn}, then 
\[\bigvee_{j=1}^m(a_{ij}^+ \ast \hat{x}_j) \vee (a_{ij}^-\ast n_P (\hat{x}_j))= b_i,\qquad i\in\{1,\dots,n\}\]
Therefore, since $\{j_1,\dots,j_l\}\subseteq\{1,\dots,m\}$ and $I_{\hat{J}^-}\subseteq\{1,\dots,n\}$, we can assert that 
\[\bigvee_{j\in\hat{J}^+}(a_{ij}^+ \ast \hat{x}_j) \vee (a_{ij}^-\ast n_P (\hat{x}_j))\leq b_i,\qquad i\in I_{\hat{J}^-}\]
Now, consider fixed $i\in I_{\hat{J}^-}$. On the one hand, if $b_i=0$, as  System~\eqref{sys:general_mxn} is solvable, then Theorem~\ref{th:sys_general_mxn} allows us to assert that $a_{ij}^+=0$ for each $j\in\hat{J}^+$. Furthermore, by definition of the mapping $f$, $\hat{x}_j>0$ for each $j\in\hat{J}^+$, and thus $n_p(\hat{x}_j)=0$. As a result, we obtain that
\[\bigvee_{j\in\hat{J}^+}(a_{ij}^+ \ast \hat{x}_j) \vee (a_{ij}^-\ast n_P (\hat{x}_j))=0\]
That is, $(\hat{x}_{j_1},\dots,\hat{x}_{j_l})$ is a solution of the $i$-th equation of System~\eqref{eq:sys_minimales}.

On the other hand, if $b_i>0$, by definition of the set $I_{\hat{J}^-}$, the inequality $a_{ij}^-<b_i$ holds for each $j\in \hat{J}^-$. Hence, due to $(\hat{x}_1,\dots,\hat{x}_m)$ is a solution of System~\eqref{sys:general_mxn}, there exists $k\in\hat{J}^+$ such that $a_{ik}^+\geq b_i$ and $a_{ik}^+*\hat{x}_k=b_i$. As a consequence, we conclude that 
\[\bigvee_{j\in\hat{J}^+}(a_{ij}^+ \ast \hat{x}_j) \vee (a_{ij}^-\ast n_P (\hat{x}_j))= b_i\]
That is, the tuple $(\hat{x}_{j_1},\dots,\hat{x}_{j_l})$ is a solution of the $i$-th equation in System~\eqref{eq:sys_minimales}. Following an analogous reasoning for each $i\in I_{\hat{J}^-}$, we conclude that $(\hat{x}_{j_1},\dots,\hat{x}_{j_l})$ is a solution of System~\eqref{eq:sys_minimales}.

Now, we are going to see that $(x_{j_1},\dots,x_{j_l})$ is also a solution of System~\eqref{eq:sys_minimales}.  Notice that, if $x_j>0$ for each $j\in\hat{J}^+$, we can repeat the previous reasoning about the tuple $(x_{j_1},\dots,x_{j_l})$. Therefore, we will see that $x_j>0$ is satisfied, for each $j\in\hat{J}^+$, by reduction to the absurd.

Suppose that $x_k=0$ for some $k\in\hat{J}^+$. As a consequence, $k\notin\hat{J}^-$. Now, defining the sets $J^+=\{j\in\{1,\dots,m\}\mid x_j>0\}$ and $J^-=\{j\in\{1,\dots,m\}\mid x_j=0\}$, we obtain that $k\in J^-$. Furthermore, by definition, $\hat{x}_j=0$ for each $j\in\hat{J}^-$. This fact together with the inequality $(x_{j_1},\dots,x_{j_l})<(\hat{x}_{j_1},\dots,\hat{x}_{j_l})$ allows us to assert that $x_j=0$ for each $j\in\hat{J}^-$, and thus $\hat{J}^-\subseteq J^-$. Hence, since $k\in J^-$ and $k\notin\hat{J}^-$, we deduce that $\hat{J}^-\subset J^-$. Finally, following the reasoning provided in the proof of Theorem~\ref{th:sys_general_mxn}, we can assert that $(J^+,J^-)$ is a feasible pair, and thus $J^-\in S^-$. As a consequence, $\hat{J}^-$ is not a maximal element in $S^-$, in contradiction with the hypothesis.

We conclude then that System~\eqref{eq:sys_minimales} has two different solutions $(x_{j_1},\dots,x_{j_l})$ and $(\hat{x}_{j_1},\dots,\hat{x}_{j_l})$, which contradicts the hypothesis. Therefore, the tuple $(\hat{x}_1,\dots,\hat{x}_m)$ is a minimal solution of System~\eqref{sys:general_mxn}, and then the mapping $f$ is well-defined. 

Finally, we will prove that $f$ is an bijective mapping from $A$ to $B$. The fact that $f$ is order-embedding is directly obtained from its definition. Let $\hat{J}^-_1,\hat{J}^-_2\in A$ with $\hat{J}^-_1\neq \hat{J}^-_2$. Without loss of generality, we can suppose that there exists $k\in \hat{J}^-_1$ such that $k\notin \hat{J}^-_2$. As a consequence, if $f(\hat{J}^-_1)=(\hat{x}_1^1,\dots,\hat{x}_m^1)$ and $f(\hat{J}^-_2)=(\hat{x}_1^2,\dots,\hat{x}_m^2)$, by definition of the mapping $f$, $\hat{x}_k^1=0$ and $\hat{x}_k^2>0$, which implies that $(\hat{x}_1^1,\dots,\hat{x}_m^1)\neq(\hat{x}_1^2,\dots,\hat{x}_m^2)$. That is, $f(\hat{J}^-_1)\neq f(\hat{J}^-_2)$, and thus $f$ is order-embedding.

To finish with this demonstration, we show that $f$ is onto. Let $(\hat{x}_1,\dots,\hat{x}_m)$ be a minimal solution of System~\eqref{sys:general_mxn}, and consider the sets
\begin{eqnarray*}
\hat{J}^+&=&\{j\in\{1,\dots,m\}\mid \hat{x}_j>0\}\\
\hat{J}^-&=&\{j\in\{1,\dots,m\}\mid \hat{x}_j=0\}
\end{eqnarray*}
Following an analogous reasoning to the proof in Theorem~\ref{th:sys_general_mxn}, we obtain that $(\hat{J}^+,\hat{J}^-)$ is a feasible pair, and thus $\hat{J}^-$ belongs to $S^-$. 

Before proving that $\hat{J}^-\in A$, we will demonstrate by reduction to the absurd that  $\hat{x}_{j}=\min\{b_h\leftarrow_P a_{hj}^+\mid h\in\{1,\dots,n\}\}$ for each $j\in\hat{J}^+$. Suppose then that there exists $k\in\hat{J}^+$ such that $\hat{x}_{k}\neq\min\{b_h\leftarrow_P a_{hk}^+\mid h\in\{1,\dots,n\}\}$. Clearly, if $\min\{b_h\leftarrow_P a_{h{k}}^+\mid h\in\{1,\dots,n\}\}<\hat{x}_{k}$, then there exists $i\in\{1,\dots,n\}$ such that $b_i\leftarrow_P a_{i{k}}^+<\hat{x}_{k}$. Applying the adjoint property, the inequality $b_i<a_{i{k}}^+*\hat{x}_{k}$ is satisfied, and therefore
\[b_i<a_{i{k}}^+*\hat{x}_{k}\leq\bigvee_{j=1}^m(a_{ij}^+ \ast \hat{x}_j) \vee (a_{ij}^-\ast n_P (\hat{x}_j))\]
That is, the tuple $(\hat{x}_1,\dots,\hat{x}_m)$ is not a solution of System~\eqref{sys:general_mxn}, in contradiction with the hypothesis.

On the contrary, assume that $\hat{x}_{k}<\min\{b_h\leftarrow_P a_{h{k}}^+\mid h\in\{1,\dots,n\}\}$. In this case, we obtain that $\hat{x}_{k}<b_h\leftarrow_P a_{h{k}}^+$ for each $h\in\{1,\dots,n\}$. Equivalently, $a_{h{k}}^+*\hat{x}_{k}<b_h$ for each $h\in\{1,\dots,n\}$. Hence, as $*$ is an order-preserving mapping, the inequality $a_{h{k}}^+*\frac{\hat{x}_{k}}{2}<b_h$ is verified, for each $h\in\{1,\dots,n\}$.

Now, according to the fact that $(\hat{x}_1,\dots,\hat{x}_m)$ is a solution of System~\eqref{sys:general_mxn}, the next expression holds:
\[\bigvee_{j=1}^m(a_{ij}^+ \ast \hat{x}_j) \vee (a_{ij}^-\ast n_P (\hat{x}_j))= b_i,\qquad i\in\{1,\dots,n\}\]
and therefore
\[\left(\bigvee_{\substack{j\in\{1,\dots,m\}\\j\neq k}}(a_{ij}^+ \ast \hat{x}_j) \vee (a_{ij}^-\ast n_P (\hat{x}_j))\right) \vee\left(a_{i{k}}^+ \ast \frac{\hat{x}_{k}}{2}\right) \vee \left(a_{i{k}}^-\ast n_P \left(\frac{\hat{x}_{k}}{2}\right)\right)= b_i\]
In other words, the tuple $(\hat{x}_1,\dots,\hat{x}_{k-1},\frac{\hat{x}_{k}}{2},\hat{x}_{k+1},\dots,\hat{x}_m)$ is a solution of System~\eqref{sys:general_mxn} which is clearly strictly smaller than $(\hat{x}_1,\dots,\hat{x}_m)$, since $\hat{x}_{k}>0$. This fact contradicts the hypothesis that $(\hat{x}_1,\dots,\hat{x}_m)$ is a minimal solution of System~\eqref{sys:general_mxn}. Hence, we ensure that $\hat{x}_j= \min\{b_h\leftarrow_P a_{hj}^+\mid h\in\{1,\dots,n\}\}$, for each $j\in \hat{J}^+$.

In the sequel, we will see that $\hat{J}^-\in A$. In order to reach this conclusion, we have to prove that $\hat{J}^-$ is a maximal element of $S^-$ and that System~\eqref{eq:sys_minimales} has only one solution. We will demonstrate these two statements by reduction to the absurd.
\begin{itemize}
\item Suppose that $\hat{J}^-$ is not a maximal element of $S^-$. In other words, suppose that there exists $J^-\in S^-$ such that $\hat{J}^-\subset J^-$. Consider the set $J^+=\{1,\dots,m\}\backslash J^-$ and the tuple $(x_1,\dots,x_m)$ given by
\[x_j= \left\{\begin{array}{lc}
             0 & \hbox{ if}\quad j\in J^-\\
             \min\{b_h\leftarrow_P a_{hj}^+\mid h\in\{1,\dots,n\}\} & \hbox{ if}\quad j\in J^+ 
             \end{array}
  \right.\]
The pair $(J^+,J^-)$ is clearly feasible, and thus following an analogous reasoning to the proof in Theorem~\ref{th:sys_general_mxn}, $(x_1,\dots,x_m)$ is a solution of System~\eqref{sys:general_mxn}. In the following, we will see that $(x_1,\dots,x_m)<(\hat{x}_1,\dots,\hat{x}_m)$, in contradiction with the hypothesis, that is $(\hat{x}_1,\dots,\hat{x}_m)$ be a minimal solution of System~\eqref{sys:general_mxn}. 

Notice that, as $\hat{J}^-\subset J^-$, by definition $x_j=\hat{x}_j=0$ for each $j\in\hat{J}^-$. Furthermore, there exists $k\in J^-$ such that $k\notin \hat{J}^-$, and thus $x_k=0<\hat{x}_k$. In addition, as $\hat{x}_{j}=\min\{b_h\leftarrow_P a_{hj}^+\mid h\in\{1,\dots,n\}\}$, for each $j\in\hat{J}^+$, then $x_j\leq\hat{x}_{j}$ holds, for each $j\in\hat{J}^+$, and hence $(x_1,\dots,x_m)<(\hat{x}_1,\dots,\hat{x}_m)$. This fact contradicts the hypothesis that $(\hat{x}_1,\dots,\hat{x}_m)$ is a minimal solution of System~\eqref{sys:general_mxn}. Hence, we conclude that $\hat{J}^-$ is a maximal element of $S^-$.

\item Suppose that System~\eqref{eq:sys_minimales} does not have a unique solution. Since the tuple $(\hat{x}_{j_1},\dots,\hat{x}_{j_l})$ is a solution of System~\eqref{eq:sys_minimales}, we obtain then that there exists another solution $(x_{j_1},\dots,x_{j_l})$ of System~\eqref{eq:sys_minimales} different from $(\hat{x}_{j_1},\dots,\hat{x}_{j_l})$. Due to $\hat{x}_{j}=\min\{b_h\leftarrow_P a_{hj}^+\mid h\in\{1,\dots,n\}\}$, for each $j\in\hat{J}^+$, we obtain that $x_j\leq\hat{x}_j$ for each $j\in\hat{J}^+$. In other words, $(x_{j_1},\dots,x_{j_l})\leq(\hat{x}_{j_1},\dots,\hat{x}_{j_l})$. Furthermore, as $(x_{j_1},\dots,x_{j_l})\neq(\hat{x}_{j_1},\dots,\hat{x}_{j_l})$, we can assert that $(x_{j_1},\dots,x_{j_l})<(\hat{x}_{j_1},\dots,\hat{x}_{j_l})$.

Hence, defining the tuple $(x^*_1,\dots,x^*_m)$ as
\[x^*_j= \left\{\begin{array}{lc}
             x_j & \hbox{ if}\quad j\in \hat{J}^+ \\
             0 & \hbox{ if}\quad j\in \hat{J}^-
             \end{array}
  \right.\]
we obtain that $(x_1,\dots,x_m)<(\hat{x}_1,\dots,\hat{x}_m)$ and it is straightforwardly a solution of System~\eqref{sys:general_mxn}. This fact contradicts the minimality of $(\hat{x}_1,\dots,\hat{x}_m)$ as a solution of System~\eqref{sys:general_mxn}. Therefore, we conclude that System~\eqref{eq:sys_minimales} has a unique solution.
\end{itemize}

We conclude then that $\hat{J}^-$ belongs to the set $A$. Clearly, according to the definition of the mapping $f$, we can assert that $f(\hat{J}^-)=(\hat{x}_1,\dots,\hat{x}_m)$. Hence, the mapping $f$ forms a bijection between $A$ and $B$.
\end{proof}

The following corollary is straightforwardly obtained from Theorem~\ref{th:sys_minimales}.

\begin{corollary}\label{cor:sys_minimales}
  Let $a_{ij}^+,a_{ij}^-,b_i,\,x_j\in[0,1]$, for each $i\in\{1,\dots,n\}$ and $j\in\{1,\dots,m\}$. Consider that System~\eqref{sys:general_mxn}
  is a solvable bipolar max-product FREs system and let $S$, $S^-$ be the sets defined as:
   \[S=\{(J^+,J^-)\mid J^+,J^-\subseteq\{1,\dots,m\}, (J^+,J^-) \hbox{ is a feasible pair w.r.t System~\eqref{sys:general_mxn}}\}\]
  \[S^-=\{J^-\mid(J^+,J^-)\in S\}\]
  Consider, for each $J^-\in S^-$, the set $I_{J^-}$ defined as:
  \[I_{J^-}=\{i\in\{1,\dots,n\}\mid b_i=0\hbox{ or }a_{ij}^-<b_i \hbox{ for each } j\in J^-\}\]
  and the mapping $f\colon S^-\to [0,1]^m$ which associates each $J^-\in S^-$ with the tuple $(x_1,\dots,x_m)$ defined, for each $j\in\{1,\dots,m\}$, by
\[x_j= \left\{\begin{array}{lc}
             0 & \hbox{ if}\quad j\in J^- \\
             \min\{b_h\leftarrow_P a_{hj}^+\mid h\in\{1,\dots,n\}\} & \hbox{ if}\quad j\notin J^-
             \end{array}
  \right.\]
  Then, the following statements hold:
  \begin{enumerate}
  \item[(1)] If $\{1,\dots,m\}\in S^-$ then $(0,\dots,0)$ is the least solution of System~\eqref{sys:general_mxn}.
  \item[(2)] If $\{1,\dots,m\}\notin S^-$,   the set of minimal solutions of System~\eqref{sys:general_mxn} is given by:
  \[\{f(J^-)\mid J^-\in M^-\}\]
  where  $M^-$ is the set of maximal elements of $S^-$ such that System~\eqref{eq:sys_minimales} has a unique solution.
  \end{enumerate}
\end{corollary}

The theoretical study carried out in this paper will be illustrated by using a toy example in the next section.

\section{A toy example}
This section will be devoted to illustrate how bipolar max-product FREs can be employed to represent a real-world situation. A toy example, in which a system of bipolar max-product fuzzy relation equations is capable of modeling the behaviour of a motor, will be presented.  Specifically, the system of bipolar max-product FREs will be employed to determine what the reasons/causes of the levels of overheating and relative humidity inside the motor are and how a technician should perform in order to the motor works properly.

\begin{example}\label{ex:motor}
Technician experts establish that a motor works in an suitable way when its temperature and its relative humidity are maintained under certain threshold, and therefore, they strongly recommend that the water level, the oil level and the functioning of radiator fan be controlled. Taking into account this information, we will model an experimental situation corresponding to the behaviour of the motor. We need to introduce the following notation in order to carry out this task. 

The water and oil levels are represented by the variables $x_1$ and $x_2$, respectively, which take values in the unit interval $[0,1]$. In particular, the value $1$ indicates that the water/oil container is empty and the value $0$ that the quantity of water/oil has  exceeded the permitted limit. The functioning of the radiator fan is represented by the variable $x_3\in\{0,1\}$, where the value $0$ means that the radiator fan is working and the value $1$ that the radiator fan is stopped. The overheating level of the motor is represented by a value $b_1\in[0,1]$, where the value $0$ indicates a correct temperature and the value $1$ a critical overheating level. The relative humidity inside the motor is represented by a value $b_2\in[0,1]$, where the value  $0$ evinces that the humidity is appropriate and the value $1$ that humidity level is critical.

Once the notation has been introduced, we can formalize the conclusions reached by technician experts on the performance of the motor. In what regards the temperature of the motor, we will assume that the motor behaves analogously to the motor in Example 19 of~\cite{CLM:JCAM2018}. In the sequel, for the sake of a self-contained example, we will remind the experts conclusions included in~\cite{CLM:JCAM2018}.

\begin{itemize}
\item The motor overheating is directly proportional to the lack of water, with proportionality constant $0.4$. Furthermore, the motor overheats at $0.7$ when there is an excess of water. Therefore, the overheating caused by the water level can be interpreted by using the expression $(0.4 \ast x_1) \vee (0.7\ast n_P (x_1))$. It is important to observe that the level of overheating is low when the water container is almost full but not exceeding the limits. This fact is due to if $x_1>0$ then $n_P (x_1)=0$. 

\item The motor overheating is also directly proportional to the lack of oil, being the proportionality constant $0.2$ in this case. In addition, an overheating of $0.1$ occurs when the oil exceeds the permitted limit. Hence, the overheating caused by the oil level can be modeled by the expression  $(0.2 \ast x_2) \vee (0.1\ast n_P (x_2))$.

\item The radiator fan does not work correctly and it sometimes suddenly stops. The motor overheats up to $0.5$ when this happens. Moreover, the standard behaviour of the radiator fan makes that the motor overheats at $0.2$. Thus, the overheating caused by the radiator fan can be interpreted by using the expression $(0.5 \ast x_3) \vee (0.2\ast n_P (x_3))$.
\end{itemize}

As it was stated in~\cite{CLM:JCAM2018}, the previous statements lead us to the following bipolar max-product fuzzy relation equation, which summarizes the effect of the water level, the oil level and the radiator fan on the temperature of the motor: 
\[(0.4 \ast x_1) \vee (0.7\ast n_P (x_1))\vee(0.2 \ast x_2) \vee (0.1\ast n_P (x_2))\vee(0.5 \ast x_3) \vee (0.2\ast n_P (x_3))=b_1\]
Concerning the relative humidity of the motor, the experts made the next assertions.

\begin{itemize}
\item The relative humidity inside the motor is not affected by the water level, since the water container is properly isolated. However, the relative humidity can increase up to 0.9 when the water exceeds the limit. This fact can be modeled by the expression $0.9\ast n_P (x_1)$.

\item The relative humidity inside the motor is not affected by the oil level. 

\item Finally, the radiator fan produces an increase in the relative humidity up to $0.4$ whenever it is stopped. As a result, the relative humidity caused by the radiator fan can be interpreted as $0.4 \ast x_3$.
\end{itemize}

Hence, in this case, the impact of the the water level, the oil level and the radiator fan on the relative humidity inside the motor can be modelled as
\[(0.9\ast n_P (x_1))\vee(0.4 \ast x_3)=b_2\]

According to the previous considerations, based on the technician experts knowledge, the reasons/causes of overheating and relative humidity level can be inferred from the system of bipolar max-product fuzzy relation equations given below:
\begin{eqnarray} \label{sys:ejemplo2}
(0.4 \ast x_1) \vee (0.7\ast n_P (x_1))\vee(0.2 \ast x_2) \vee (0.1\ast n_P (x_2))\vee(0.5 \ast x_3) \vee (0.2\ast n_P (x_3))=b_1\\
(0.9\ast n_P (x_1))\vee(0.4 \ast x_3)=b_2\nonumber
 \end{eqnarray}
 
Now, we will suppose that the motor presents an overheating of $b_1=0.3$ but its relative humidity is under control, that is, $b_2=0$. It would be interesting to know what values of water, oil and radiator fan are producing this performance of the motor. In the sequel, we will see that the previous system provides a useful tool in order to determine these values.

Defining $J_1^+=\{1,2\}$ and $J_1^-=\{3\}$, one can easily check that $(J_1^+,J_1^-)$ forms a feasible pair w.r.t. System~\eqref{sys:ejemplo2}. As a result, Theorem~\ref{th:sys_general_mxn} allows us to assert that System~\eqref{sys:ejemplo2} is solvable. Furthermore, there are only two feasible pairs w.r.t. System~\eqref{sys:ejemplo2} which are $(J_1^+,J_1^-)=(\{1,2\},\{3\})$ and $(J_2^+,J_2^-)=(\{1\},\{2,3\})$  

On the one hand, according to Theorem~\ref{th:sys_maximales}, since $J_1^+$ is the greatest element of the set $S^+=\big\{\{1,2\},\{1\}\big\}$, we conclude that there exists the greatest solution of System~\eqref{sys:ejemplo2}. Specifically, it is given by the tuple $(0.75,1,0)$.

On the other hand, concerning to the existence of minimal solutions, we obtain that $S^-=\big\{\{3\},\{2,3\}\big\}$, and thus $\{1,2,3\}\notin S^-$. Notice that, $J^-_2$ is the unique maximal element of the set $S^-$. In addition, since $b_2=0$ and the inequalities $a_{12}^-=0.1<0.3=b_1$ and $a_{13}^-=0.2<0.3=b_1$ hold, the set $I_{J_2^-}$ is defined as:
\[I_{J_2^-}=\{i\in\{1,\dots,n\}\mid b_i=0\hbox{ or }a_{ij}^-<b_i \hbox{ for each } j\in J_2^-\}=\{1,2\}\]
Consequently, System~\eqref{eq:sys_minimales} of Theorem~\ref{th:sys_minimales} is given by:
\begin{eqnarray*}
	(0.4\ast x_1) \vee (0.7\ast n_P (x_1))=b_1\\
	(0.9\ast n_P (x_1))=b_2
  \end{eqnarray*}
whose unique solution is $x_1=0.75$. Hence, Theorem~\ref{th:sys_minimales} leads us to conclude that System~\eqref{sys:ejemplo2} has only one minimal solution. In particular, $(0.75,0,0)$ is the minimal solution of System~\eqref{sys:ejemplo2}.

According to the greatest solution and the minimal solution of System~\eqref{sys:ejemplo2}, we deduce then that the water level is equal to $0.75$ and that the radiator fan is properly working, since its value is $0$ in both cases. Notice that, the oil level can take any value in $[0,1]$ and therefore we cannot infer any information from the oil level. Nevertheless, in this particular situation, we can ignore the oil level since in the worst case it can gives rise an overheating of $0.2$ and it does not affect to the relative humidity.

From all this information, we would suggest to a technician refilling the water container, being careful in order to do not exceed the limit.
\qed
\end{example}

We have included a toy example in order to complement our study exposing the possibilities of this contribution to practical applications.  Specifically, the situation and the variables involved in the behaviour of the motor have been presented. Afterwards, the existing relations between the variables have been detailed and we have interpreted these relations as a system of bipolar max-product fuzzy relations equations with the product negation. In this procedure, one can realise that the product operator becomes crucial for modelling a proportional relation between two variables. Besides, the usefulness of the negation operator $n_P$ in what regards to interpreting interruptions in a  continuous relation have been shown. For instance, the sudden changes in the motor overheating when the level of water or oil exceeds the permitted limit are modelled through the operator $n_P$.

Once the system of bipolar max-product fuzzy relations equations with the product negation has been defined, the main results of the paper have been applied. In particular, we make use of Theorems~\ref{th:sys_general_mxn}, \ref{th:sys_maximales} and~\ref{th:sys_minimales}, which concern the solvability and the algebraic structure of the set of solutions of System~\eqref{sys:ejemplo2}, in order to infer outcomes from a given real-world situation. 

\section{Conclusions and future work}
We have solved one open problem given in~\cite{CLM:JCAM2018}, introducing a characterization on the resolution of any bipolar max-product FREs system, including the systems with several zero elements in the independent terms. We have complemented this characterization with diverse  properties related to the algebraic structure of the set of solutions of these systems. Hence, this paper has completed the study on the resolution of any kind of fuzzy relation equation with the product t-norm and its residuated negation.    This t-norm is one of the most important and useful operators, which is very natural in the applications, 
as we have shown in Example~\ref{ex:motor}.

The consideration of arbitrary negations and the use of other general operators, such as uninorms and u-norms, instead of the product t-norm,  will be fundamental in the development of future advances in this research topic.

\end{document}